\documentclass[10pt,journal,compsoc]{IEEEtran}

\ifCLASSOPTIONcompsoc
  \usepackage[nocompress]{cite}
\else
  \usepackage{cite}
\fi

\ifCLASSINFOpdf
\else
\fi

\usepackage{graphicx}
\usepackage[cmex10]{amsmath}
\usepackage{subfig}
\usepackage{array}
\usepackage{booktabs}
\usepackage{multirow}
\usepackage{threeparttable}
\usepackage{amsfonts}
\usepackage{tabularx}
\usepackage{algorithm}
\usepackage{algorithmic}
\usepackage{amsmath}
\usepackage{amssymb}
\DeclareMathOperator*{\argmax}{arg\,max}

\usepackage{mathtools}
\usepackage{cuted}

\newcolumntype{Y}{>{\centering\arraybackslash}X}
\newcolumntype{b}{>{\centering\arraybackslash}X}
\newcolumntype{s}{>{\hsize=.33\hsize\centering\arraybackslash}X}

\makeatletter
\newcommand*{\rom}[1]{\expandafter\@slowromancap\romannumeral #1@}
\makeatother

\begin{document}
%
\title{Neural Belief Propagation for Scene Graph Generation}
%
%
%
%

\author{Daqi~Liu,~Miroslaw~Bober,~\IEEEmembership{Member,~IEEE},~Josef~Kittler,~\IEEEmembership{Life Member,~IEEE}
\IEEEcompsocitemizethanks{\IEEEcompsocthanksitem The authors are with the Centre for Vision, Speech and Signal Processing, University of Surrey, Guildford GU2 7XH, U.K. 
\textbf{Under Review...} \protect\\
E-mail: \{daqi.liu, m.bober, j.kittler\}@surrey.ac.uk}
}

%

\IEEEtitleabstractindextext{%
\begin{abstract}
Scene graph generation aims to interpret an input image by explicitly modelling the potential objects and their relationships, which is predominantly solved by the message passing neural network models in previous methods. Currently, such approximation models generally assume the output variables are totally independent and thus ignore the informative structural higher-order interactions. This could lead to the inconsistent interpretations for an input image. In this paper, we propose a novel neural belief propagation method to generate the resulting scene graph. It employs a structural Bethe approximation rather than the mean field approximation to infer the associated marginals. To find a better bias-variance trade-off, the proposed model not only incorporates pairwise interactions but also higher order interactions into the associated scoring function. It achieves the state-of-the-art performance on various popular scene graph generation benchmarks. 

\end{abstract}

}

\maketitle

\IEEEdisplaynontitleabstractindextext

%
\IEEEpeerreviewmaketitle

\IEEEraisesectionheading{\section{Introduction}\label{sec:introduction}}

%
%
%
%
\IEEEPARstart{S}{cene} graph generation (SGG) is a type of structured prediction task, which, given an input image, explicitly models the objects appearing in the scene and their relationships. Generally, such a structured prediction task can be modeled by discriminative undirected probabilistic graphical models, i.e. conditional random fields (CRFs) \cite{sutton2006introduction}, \cite{zheng2015conditional}. Given an input image $I$, SGG aims to infer the optimum interpretation $x^*$ via a maximization procedure $x^*=argmax_{x}S(I,x)$, where $S(I,x)$ represents a scoring function, which measures the consistency between the input image $I$ and the output interpretation $x$. Such an inference task is generally NP-hard to solve, due to the combinatorial nature of the structured outputs in SGG applications. To this end, approximation strategies such as mean field variational inference \cite{wainwright2008graphical}, \cite{fox2012tutorial} or loopy belief propagation \cite{murphy1999loopy}, \cite{ihler2005loopy} are required to address the above NP-hard inference task.

Current SGG models generally follow a unique paradigm, which consists of two fundamental modules: visual perception and visual context reasoning \cite{liu2019visual}. The former extracts a set of region proposals within the input image while the latter infers the optimum instance/relationship interpretations for those region proposals. Within the visual context reasoning module, the corresponding scoring function is generally computed as:
\begin{equation}
S(I,x)=\prod_{r\in R}f_{r}(I, x_r)
\end{equation}
where $r$ is a clique (a subset of vertices of an undirected graph) within a clique set $R$, $f_r$ is a non-negative function which measures the similarities between the input image $I$ and the associated clique variables $x_r$. In current SGG models, only two types of cliques are considered: 1-vertex cliques (for unary interactions) and 2-vertex cliques (for pairwise interactions), as demonstrated in Fig.1. 

With the above scoring function, one can directly compute the model posterior $p(x|I)$ as:
\begin{equation}
p(x|I)=\frac{S(I,x)}{\sum_{x}S(I,x)}=\frac{S(I,x)}{S(I)}
\end{equation}
where  $S(I)$ is the partition function. Due to the exponential structural outputs in SGG tasks, such model posterior $p(x|I)$ is generally computationally intractable, and thus it is often approximated by a computationally tractable variational distribution $q(x)$. From a broad perspective, the above paradigm can also be interpreted as a variational Bayesian \cite{wainwright2008graphical}, \cite{fox2012tutorial} framework.

To construct a variational approximation $q(x)$ for the model posterior $p(x|I)$, in current SGG models, message passing neural network becomes a universal solution, which generally assumes the output variables are totally independent:
\begin{equation}
q(x)=\prod_{i=1}^{n}q_i(x_i)
\end{equation}
where $q_i(x_i)$ represents the local variational approximation of the $i$-th output variable. The informative higher-order (including pairwise) interactions among the output variables are largely ignored in the above decomposition. From this perspective, the above ubiquitous formulation can be considered as a neural mean field variational inference method. 	

However, the above structural higher order interactions play an important role in generating coherent interpretations, which could be explained from the following two aspects: 1) the structural higher-order interactions act as a regularizer for the applied variational approximations and thus reduce the search space; 2) the variaitional distribution with the higher order interactions is a tighter approximation of the model posterior. Moreover, the scoring functions applied in current SGG models only consider unary and pairwise interactions, which may not capture the complex dependencies existing among the output variables within the SGG systems. 

To address the above issues, inspired by the recently proposed factor graph neural network (FGNN) model \cite{zhang2020factor}, we propose a novel neural belief (NBP) propagation method, which employs a structural Bethe approximation \cite{yedidia2003understanding}, \cite{yedidia2005constructing} rather than the naive mean field approximation \cite{wainwright2008graphical}, \cite{fox2012tutorial} to infer the associated marginals. Specifically, to capture the global structural interactions among the output variables, a new scoring function is defined in this paper, which extends the current pairwise interactions to even higher order interactions. This essentially reduces the model bias so that we could find a better bias-variance trade-off. More importantly, based on the FGNN architecture, we propose a novel neural belief propagation framework, which essentially aims to simulate a classical sum-product algorithm \cite{kschischang2001factor}. The proposed generic NBP method is validated on two popular SGG benchmarks: Visual Genome and Open Images V6. It achieves the state-of-the-art performance.

This paper is organized as follows: Section 2 and Section 3 present the background and the related work, respectively. Section 4 introduces the proposed neural belief propagation method. The experimental results and the associated analysis are elaborated in Section 5. Finally, the conclusions are drawn in Section 6.

\section{Background}

\subsection{Factor Graph}

Factor graph \cite{wainwright2008graphical}, \cite{loeliger2007factor} is a bipartite probabilistic graphical model, which aims to model the dependencies among the random variables via factorizing a corresponding scoring function:
\begin{equation}
S(x)=\prod_{r\in R}f_r(x_r)
\end{equation}
where the dependencies among each subset of variables (clique) $x_r$ are modeled by a corresponding non-negative factor function $f_r$. Fig.1 demonstrates a simple factor graph, in which factor $f_1$ depends on subset $\{x_1\}$, $f_2$ depends on subset $\{x_1, x_3\}$ and $f_3$ depends on subset $\{x_2,x_3\}$. Here, $f_1$ corresponds to a 1-vertex clique, while $f_2$ and $f_3$ have 2-vertex cliques. With the above scoring function $S(x)$, one can compute the corresponding probability distribution $p(x)=\prod_{r\in R}f_r(x_r)/Z$, where $Z$ is the associated partition function.

\subsection{Belief Propagation}

As a message passing method, belief propagation \cite{yedidia2003understanding}, \cite{yedidia2005constructing} performs inference on graphical models by locally marginalizing over random variables, which is also known as the sum-product algorithm. Given a factor graph, belief propagation can efficiently compute the associated marginals via exploring its unique structure. Specifically, it works by sending real-valued functions called messages along the associated edges. With such messages, the nodes can exchange their beliefs about others and thus transporting the associated probabilities. 
\begin{figure}[!t]
\centering
\includegraphics[width=1.36 in]{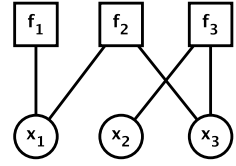}
\caption{ An example factor graph. $f_1$ corresponds to 1-vertex (subset with only one vertex variable $\{x_1\}$) clique, while $f_2$ and $f_3$ have 2-vertex (subsets with two vertex variables $\{x_1, x_3\}$ and $\{x_2, x_3\}$, respectively) cliques. }
\label{fig_1}
\end{figure} 

Based on whether the node receiving the message is a variable node or factor node, there are two types of messages: Variable-to-Factor message and Factor-to-Variable message. Specifically, Variable-to-Factor message $\mu_{x_i\rightarrow f_m}$ is the product of the messages from all other neighboring factor nodes $N(x_i)$ except $f_m$:
\begin{equation}
\mu_{x_i\rightarrow f_m}=\prod_{n\in N(x_i)\backslash  f_m}\mu_{f_n\rightarrow x_i}(x_i)
\end{equation}
while Factor-to-Variable message is the product of the factor with messages from all other nodes, marginalized over all variables $x_m$ except $x_j$:
\begin{equation}
\mu_{f_m\rightarrow x_j}=\sum_{x_m \backslash x_j}f_m(x_m)\prod_{i\in N(f_m)\backslash  j}\mu_{x_i\rightarrow f_m}(x_i)
\end{equation}
where, after recursively running the above two steps until convergence, the associated marginal $p(x_j)$ can be computed as:
\begin{equation}
p(x_j)\propto \prod_{m\in N(x_j)}\mu_{f_m\rightarrow x_j}(x_j)
\end{equation}
\section{Related Work }

As a structured prediction task, scene graph generation has already become a hot topic in the computer vision area as a means to facilitate the downstream vision tasks like image captioning \cite{you2016image}, \cite{rennie2017self}, \cite{yang2019auto} and visual question answering \cite{teney2017graph}, \cite{anderson2018bottom}, \cite{shi2019explainable}. Specifically, current SGG models often aim to achieve two main objectives: extracting informative feature representations and implementing unbiased training. 
\begin{figure*}[!t]
\centering
\includegraphics[width=\linewidth]{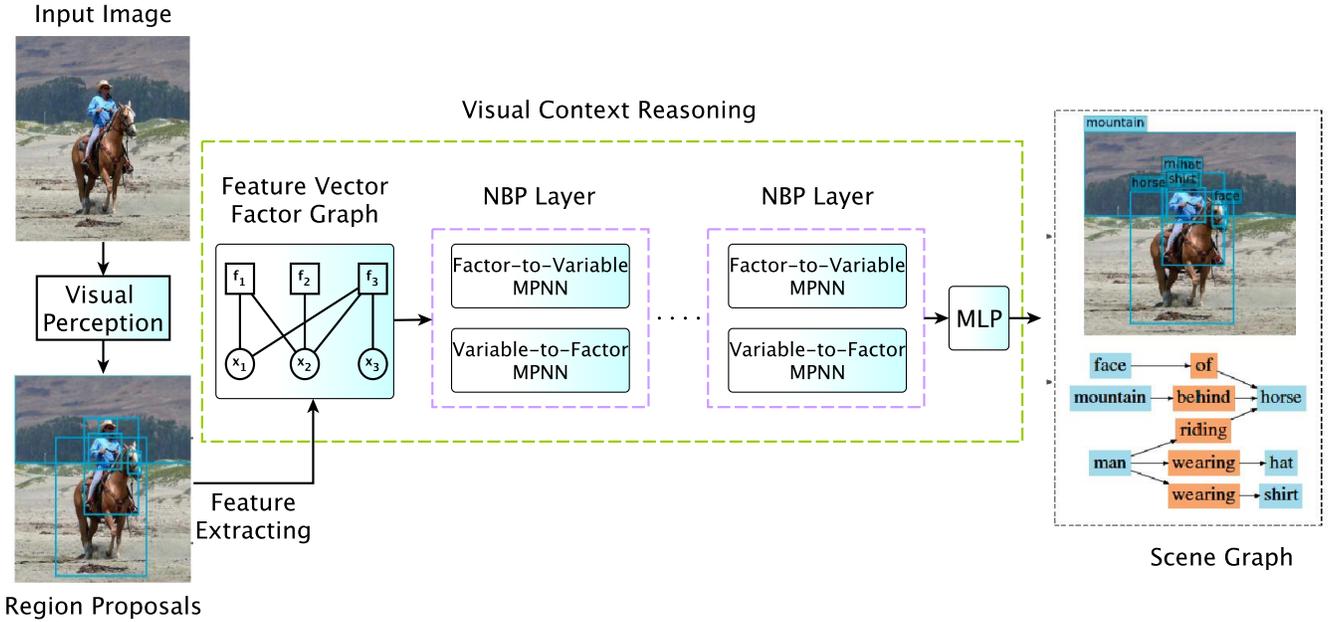}
\caption{Overview of the proposed neural belief propagation (NBP) method, in which the green dash line represents the proposed visual context reasoning module. Given an input image, visual perception module detects a set of region proposals. A feature vector factor graph $G=(X,F,E)$ is constructed based on the above region proposals, in which the vertex variable set $X$, the factor set $F$ and the related edge set $E$ are all associated with the corresponding feature vectors. With such a feature vector factor graph, one can infer the resulting marginals via stacking various NBP layers plus a final MLP. Each of the NBP layers consists of two message passing neural network (MPNN) models: Factor-to-Variable MPNN and Variable-to-Factor MPNN, which essentially correspond to the two types of messages within the classical belief propagation method. Given the resulting marginals of the factor graph $G$, one can easily compute the optimum interpretations (for related vertex variables and pairwise factors, which corresponds to instances and predicates in  a scene graph) via simple $argmax$ operations.  A cross-entropy loss is applied to train the proposed NBP method.}
\label{fig_2}
\end{figure*} 

To produce the informative feature representations from the input image, current SGG models generally follow two main directions: 1) design new message passing neural network structures \cite{yang2019auto}, \cite{dai2017detecting}, \cite{li2018factorizable}, \cite{qi2019attentive}, \cite{zellers2018neural}; 2) incorporate related contextual structural information into the existing neural network structures by investigating different message passing mechanisms \cite{qi2019attentive}, \cite{yang2018graph}, \cite{tang2019learning}, \cite{woo2018linknet}, \cite{lin2020gps}.

To alleviate the ubiquitous biased relationship prediction problem caused by the long-tail data distribution, current SGG models often rely on certain rebalancing strategies, which include dataset resampling \cite{chawla2002smote}, \cite{shen2016relay}, \cite{mahajan2018exploring}, instance-level resampling \cite{gupta2019lvis}, \cite{hu2020learning}, bi-level data resampling \cite{li2021bipartite}, loss reweighting based on instance frequency \cite{cao2019learning}, \cite{cui2019class} and knowledge transfer learning \cite{gidaris2018dynamic}, \cite{zhou2020bbn}, \cite{guo2021general}. Recently, \cite{tang2020unbiased} formulates the SGG task as a causal model and present an unbiased learning method based on causal inference. 

Besides the above mean field message passing neural networks, various papers \cite{zhang2020factor}, \cite{kuck2020belief}, \cite{satorras2021neural} were proposed to extend the current message passing neural network structures to factor graphs, which could simulate or improve the classical belief propagation methods. Such advanced message passing neural network structures break the independence assumption made by the previous models, which provide a solid basis for our proposed neural belief propagation method. Compared with the previous SGG models, the proposed NBP method has two main advantages: 1) incorporating higher order interactions into the proposed scoring function aiming to find a better bias-variance trade-off; 2) applying Bethe approximation (implemented by belief propagation algorithm) rather than the ubiquitous mean field approximation to infer the associated marginals since the latter often underestimates the underlying model posterior. 

\section{Proposed Methodology}

In this section, we first formulate the SGG task as a marginal inference model and present a novel scoring function for such a model, followed by the introduction of the proposed neural belief propagation method. Fig.2 demonstrates the overview of the proposed neural belief propagation method.  

\subsection{Problem Formulation}

As a structured prediction task, given an input image, SGG aims to model the potential objects as well as their relationships via certain inference strategies, which can be naturally modelled by a discriminative undirected probabilistic graphical model, i.e. Conditional Random Field. Such undirected graphical model can be further transformed into a corresponding factor graph, in which the classical belief propagation method can be applied to infer the associated marginal distributions.

Given an input image $I$, SGG aims to infer the optimum interpretations $x^*$ as:
\begin{equation}
x^*=\argmax_{x}p(x|I)=\argmax_{x}S(I,x)
\end{equation}
where $p(x|I)$ represents the model posterior and $S(I, x)$ denotes the associated scoring function. The above inference task is essentially formulated as a maximum aposteriori estimation problem (MAP), which is generally NP-hard to solve for structured prediction tasks like SGG.  Following \cite{meltzer2009convergent}, such NP-hard MAP inference can be formulated as an integer linear program and further transformed into a relevant relaxed linear program. More importantly, one can unify the above MAP inference with the related marginal inference as follows: 
\begin{equation}
q^*=\argmax_{q}\mathbb{E}_{q(x)}S(I,x)+T\mathbb{H}(q(x))
\end{equation}
where $q(x)$ is a variational distribution and $\mathbb{H}(q(x))$ is its entropy.  $T$ is a temperature parameter where $T=1$ for marginal inference and $T=0$ for MAP inference. In SGG tasks, to avoid underestimating the model posterior, the above entropy term is often required (where $T$ is often set to $1$). In other words, for SGG tasks, instead of solving the original MAP inference, we prefer to first infer the marginals and then compute the optimum interpretations $x^*$ via additional $argmax$ operations. Empirically, we find the above solution often produces a better performance than the exact MAP inference route (implemented by the max-product NBP). 

Moreover, the factorization of the above scoring function is generally formulated as a bipartite factor graph. Assuming the scoring function $S(I,x)$ of an SGG task can be decomposed as:
\begin{equation}
S(I,x_1, x_2, ..., x_n)=\prod_{j}^{m}f_{j}(I, C_j)
\end{equation}
where $C_j \subseteq \{x_1, x_2, ..., x_n\}$ is a potential clique and $f_j$ represents a non-negative factor function which characterizes the interactions among the vertex variables within $C_j$, its corresponding factor graph is represented as $G=(X,F,E)$, in which $X$ represents a variable vertex set including all the potential instances $(x_1, x_2, ..., x_n)$ detected by the associated visual perception module, $F$ is a corresponding factor set containing all the potential relationships $(f_1, f_2, ..., f_m)$ among the instances, $E$ is an edge set in which the edge $e_{ij}$ only exists if the factor $f_j$ is connected to the variable vertex $x_i$. 

Given an instance interpretation set $\mathcal{C}$ and a relationship interpretation set $\mathcal{R}$, an SGG task can be modelled as a corresponding genetic factor graph $G=(X, F, E)$, in which one needs to infer each marginal distribution corresponding to each vertex variable $x_i \in \mathcal{C}, i=1,2,...,n$ within $X$ as well as each factor $f_j \in \mathcal{R}, j=1,2,...,m$ within $F$. Currently, only pairwise interactions (2-vertex cliques) are required to be scored in SGG tasks. More importantly, with the above factor graph $G=(X, F, E)$, one can efficiently infer the associated marginal distributions via the classical sum-product method and compute $x^*$ via additional $argmax$ operations.

\subsection{Proposed Scoring Function}

In current SGG models, the scoring function $S(I,x)$ only incorporates two types of cliques: 1-vertex cliques $C=\{x_i\},i=1,2,...,n$ and 2-vertex cliques $C=\{x_i, x_j\},j\in N(i)$, where $j$ is the vertex around the target vertex $i$. The former measures the similarity between $I$ and each potential variable vertex $x_i$ while the latter characterizes the dependency between two interacting variable vertexes $\{x_i,x_j\}$.
\begin{equation}
S(I,x)=\prod_{i}^{n}[f_{i}(I, x_i)\prod_{j \in N(i)}f_{ij}(x_j,x_i)]
\end{equation}
Such scoring function formulation only considers the pairwise interactions among the vertex variables, which may underestimate the ground-truth posterior $p_r(x|I)$. As a result, the model posterior $p(x|I)$ produced by such a scoring function is not a tight approximation of the ground-truth posterior $p_r(x|I)$. The consequence of the approximation error is a model bias.

To lower the above model bias and find a better bias-variance trade-off, in this paper, we propose a novel scoring function formulation, which incorporates multi-vertex (higher than 2-vertex) cliques into the associated scoring function. 
\begin{equation}
S(I,x)=\prod_{i}^{n}[f_{i}(I, x_i)\prod_{j \in N(i)}f_{ij}(x_j,x_i)]\prod_{h}f_h(x_{h})
\end{equation}
where $h\subseteq \{x_1, x_2, ..., x_n\}$ is a multi-vertex clique with $f_h$ as its corresponding factor function. In the proposed scoring function formulation, for an input image $I$ with $n$ detected vertex variables $\{x_1, x_2, ..., x_n\}$, the multi-vertex clique $h=\{x_1, x_2, ..., x_n\}$ includes all the potential detected vertex variables. Such formulation considers the global contextual information via incorporating the above multi-vertex cliques into the target scoring function. Essentially, it lowers the model bias by aiming to pursue a better bias-variance trade-off. Traditionally, improving the complexity of a model would undoubtedly increase the associated computational burden. To this end, a novel neural belief propagation method is introduced in the following subsection. 

\subsection{Neural Belief Propagation }

To efficiently infer the associated marginal distributions for the above complex scoring function, in this section, we propose a novel neural belief propagation (NBP) method, which combines the powers of both classical belief propagation algorithm and the modern message passing neural network structures. The proposed NBP method extends the current message passing neural network so that it can break the previous universal yet naive independence assumption. To better illustrate the proposed method, we first present a generic message passing neural network framework, followed by the introduction of a specific FGNN structure. Finally, we explain how we build the proposed NBP method based on such an FGNN structure. 

\subsubsection{Generic Message Passing Neural Network}

Following \cite{gilmer2017neural}, in this section, we present a generic message passing neural network (MPNN) structure for the current SGG models. With such generic framework, one can easily define a new graph neural network model by modifying the related message passing operations. 

Specifically, given a graph $G=(X,E)$ with a set of nodes $X$ and a set of pairwise edges $E$, suppose each node $x_i$ is associated with a feature representation $v_i$ and each possible pairwise edge is associated with a feature representation $e_{ij}$ (where $j\in N(i)$ is a neighbouring node to $i$), the above generic message passing neural network can be described as:
\begin{equation}
m_i=\sum_{j\in N(i)}M(v_i,v_j,e_{ij}),\;
\hat{v}_i=U(v_i,m_i)
\end{equation}
where $M$ and $U$ are implemented by neural networks. The summation aggregator in Equation (13) can be replaced by other operations. Such a generic framework can be generally applied to any graph with only pairwise edges.

\subsubsection{Factor Graph Neural Network}

To extend the above formulation to the generic factor graphs with extra edges, other than the pairwise ones, a factor graph neural network (FGNN) structure \cite{zhang2020factor} has recently been proposed. Specifically, FGNN follows a unique MPNN structure, which can be denoted as follows:
\begin{equation}
\hat{v}_i=\max_{j\in N(i)}Q(e_{ij})M(v_i,v_j)
\end{equation}
where $Q$ qualifies $e_{ij}$ by an $m\times n$ weight matrix and $M$ maps the associated feature vectors $[v_i,v_j]$ into a length-$n$ feature vector. Correspondingly, a new length-$m$ feature vector $\hat{v}_i$ is produced after the above matrix multiplication and aggregation. 

With the above unique MPNN structure, FGNN encodes the higher order features via incorporating extra factor nodes. Consider a factor graph $G=(X, F, E)$. Suppose each vertex variable $x_i\in X$ is associated with a feature variable $v_i$, each factor $f_j \in F$ is associated with a factor feature $g_j$, and each edge $e_{ij},j\in N(i)$ connecting $x_i$ and $f_j$ is associated with an edge representation $t_{ij}$. An FGNN layer consists of two important modules: Variable-to-Factor MPNN and Factor-to-Variable MPNN, which are defined  as follows: 
\begin{equation}
\begin{split}
\hat{g}_j=\max_{i\in N(j)}Q(t_{ij}|\phi_{VF})M([v_i,g_j]|\psi_{VF})\\
\hat{v}_i=\max_{j\in N(i)}Q(t_{ij}|\phi_{FV})M([v_i,g_j]|\psi_{FV})
\end{split}
\end{equation}
where $\phi$ and $\psi$ are used to parameterize the neural networks $Q$ and $M$, respectively. As shown in the above equation, those two MPNN modules have similar structures but different parameters. More importantly, one can simulate $k$ max-product iterations via stacking $k$ FGNN layers, plus a linear layer at the end. In other words, with the above FGNN structure, one can perform inference over the associated factor graph akin to the classical max-product method. 

\subsubsection{Neural Belief Propagation Method}

Based on the above FGNN structure, in this paper, we propose a novel neural belief propagation method for SGG tasks. Unlike the FGNN structure, the proposed NBP method aims to simulate a sum-product method since we find it generally achieves a better performance than its counterpart, the max-product version. Accordingly, the aggregator in the proposed NBP method is set to the mean operation rather than the maximum operation. 
\begin{figure}[!t]
\centering
\includegraphics[width=\linewidth]{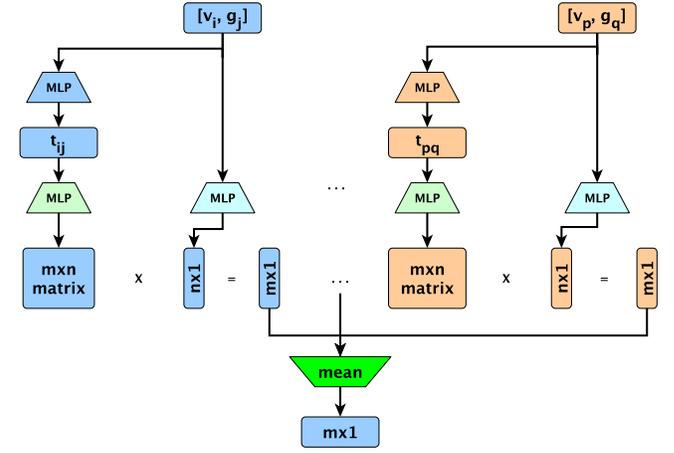}
\caption{ Illustration of the message passing neural network (MPNN) structure applied in the proposed NBP method. For each potential edge, a $m\times 1$ feature vector is produced by multiplying a corresponding $m\times n$ matrix with an associated $n\times 1$ feature vector. A mean aggregator is applied to aggregate the candidate feature vector group into a resulting $m\times 1$ feature vector.}
\label{fig_3}
\end{figure} 

Specifically, as demonstrated in Equation (12), the proposed scoring function has two types of higher order cliques: two-vertex cliques and multi-vertex cliques. To efficiently infer the marginals from the above complex scoring function, the proposed NBP method incorporates two lists of factor nodes, in which the first factor set corresponds to the two-vertex cliques while the multi-vertex cliques are modeled in the second factor set. For each factor list, two index matrices are applied in the proposed method to represent the Factor-to-Variable and Variable-to-Factor connection configurations, which essentially defines the structures for the associated Factor-to-Variable and Variable-to-Factor MPNNs. 

Following FGNN, we apply similar Variable-to-Factor and Factor-to-Variable MPNNs in the proposed NBP method. Given a feature variable $v_i$ and a factor feature $g_j$, we apply a MLP to extract the corresponding edge feature $t_{ij}$, which is further transformed into an $m\times n$ matrix via another MLP. The feature pairs $[v_i,g_j]$ are mapped into an $m\times 1$ feature vector via an associated MLP. Furthermore, a $m\times 1$ feature vector is obtained via a matrix multiplication for each potential edge. Finally, we apply a mean aggreagator to produce the resulting $m\times 1$ feature vector from the candidate feature vector group produced by the above matrix multiplications. The specific MPNN structure applied in the proposed NBP method is illustrated in Fig.3. 

With the above MPNN module, for each vertex variable (instance) and each pairwise factor variable (predicate) in $G=(X,F,E)$, an $m\times 1$ feature vector is produced, which is further mapped into a corresponding logit via an associated MLP. A cross entropy loss is employed in this paper to train the above NBP method. As a result, the proposed NBP method extends the current message passing neural networks, in which the pairwise interactions are incorporated into the associated variational approximation. In other words, one can apply a structural Bethe approximation \cite{yedidia2003understanding}, \cite{yedidia2005constructing} to replace the previous ubiquitous mean field approximation. By virtue of the proposed NBP method, a tighter variational approximation is obtained for the underlying model posterior $p(x|I)$ \cite{yedidia2003understanding}, \cite{yedidia2005constructing}.

\section{Experiments}

To validate the proposed neural belief propagation method, two popular scene graph generation benchmarks - Visual Genome \cite{krishna2017visual} and Open Images V6 \cite{alina2020open} - are utilized in this section. For each benchmark, we first introduce the experimental configuration, followed by the comparisons with the state-of-the-art methods. The ablation study and the visualization results are also included in the last two subsections.
\begin{table*}[t]
   \begin{threeparttable}
	\renewcommand{\arraystretch}{1.5}
	\caption{A performance comparison on Visual Genome dataset.}
    \centering
    \begin{tabular*}{\linewidth}{c|@{\extracolsep{\fill}}cccccc|ccc}
	\toprule
	\multirow{2}{*}{Method} & \multicolumn{2}{c}{PredCls} & \multicolumn{2}{c}{SGCls} & \multicolumn{2}{c|}{SGDet} & \multicolumn{3}{c}{SGDet(R@100)}\\ \cmidrule{2-3} \cmidrule{4-5} \cmidrule{6-7} \cmidrule{8-10} 
	 {} & {mR@50} & {mR@100} & {mR@50} & {mR@100} & {mR@50} & {mR@100} & {Head} & {Body} & {Tail}\\
	\midrule
    RelDN$ ^{\dagger}$\cite{zhang2019vrd}  & $15.8$ & $17.2$ & $9.3$ & $9.6$ & $6.0$ & $7.3$ & $34.1$ & $6.6$ & $1.1$ \\
	Motifs\cite{zellers2018neural}  & $14.6$ & $15.8$ & $8.0$ & $8.5$ & $5.5$ & $6.8$ & $36.1$ & $7.0$ & $0.0$ \\
	Motifs*\cite{zellers2018neural}  & $18.5$ & $20.0$ & $11.1$ & $11.8$ & $8.2$ & $9.7$ & $34.2$ & $8.6$ & $2.1$\\
	G-RCNN$ ^{\dagger}$\cite{yang2018graph}   & $16.4$ & $17.2$ & $9.0$ & $9.5$ & $5.8$ & $6.6$ & $28.6$ & $6.5$ & $0.1$ \\
	MSDN$ ^{\dagger}$\cite{li2017scene}   & $15.9$ & $17.5$ & $9.3$ & $9.7$ & $6.1$ & $7.2$ & $35.1$ & $5.5$ & $0.0$ \\
    GPS-Net$ ^{\dagger}$\cite{lin2020gps}   & $15.2$ & $16.6$ & $8.5$ & $9.1$ & $6.7$ & $8.6$& $34.5$ & $7.0$ & $1.0$ \\
    GPS-Net$ ^{\dagger *}$\cite{lin2020gps}   & $19.2$ & $21.4$ & $11.7$ & $12.5$ & $7.4$ & $9.5$& $30.4$ & $8.5$ & $3.8$ \\
    VCTree-TDE\cite{tang2020unbiased}  & $25.4$ & $28.7$ & $12.2$ & $14.0$ & $9.3$ & $11.1$ & $24.5$ & $13.9$ & $0.1$ \\
    BGNN\cite{li2021bipartite}  & $30.4$ & $32.9$ & $14.3$ & $16.5$ & $10.7$ & $12.6$& $33.4$ & $13.4$ & $6.4$ \\
    \textbf{NBP} & $\mathbf{28.5}$ & $\mathbf{30.6}$ & $\mathbf{15.1}$ & $\mathbf{16.5}$ & $\mathbf{12.9}$ & $\mathbf{14.7}$& $\mathbf{31.7}$ & $\mathbf{15.0}$ & $\mathbf{8.9}$ \\
	\bottomrule
    \end{tabular*}
    \begin{tablenotes}
	\item [\textbullet] Note: All the above methods apply ResNeXt-101-FPN as the backbone. $*$ means the re-sampling strategy \cite{gupta2019lvis} is applied in this method, and $\dagger$ depicts the reproduced results with the latest code from the authors. 
      \end{tablenotes}
    \end{threeparttable}
\end{table*} 

\subsection{Visual Genome}
\subsubsection{Experimental Configuration}

\textbf{Benchmark:} As the most popular SGG benchmark, Visual Genome \cite{krishna2017visual}  consists of 108,077 images with an average of 38 objects and 22 relationships per image. Following the data split protocol in \cite{xu2017scene}, in this experiment, we choose the most frequent 150 object classes and 50 predicate classes. Specifically, we split Visual Genome benchmark into two sets: a training set ($70\%$) and a test set ($30\%$). An evaluation set ($5k$) is further extracted from the training set for model validation. To investigate the biased relationship prediction problem caused by long-tail data distribution, according to the instance number in training set \cite{liu2019large}, we split the categories into three disjoint sets: $head$ (more than $10k$), $body$ ($0.5k\sim 10k$) and $tail$ (less than $0.5k$), as demonstrated in Fig.4. 
\begin{figure}[!t]
\centering
\includegraphics[width=\linewidth]{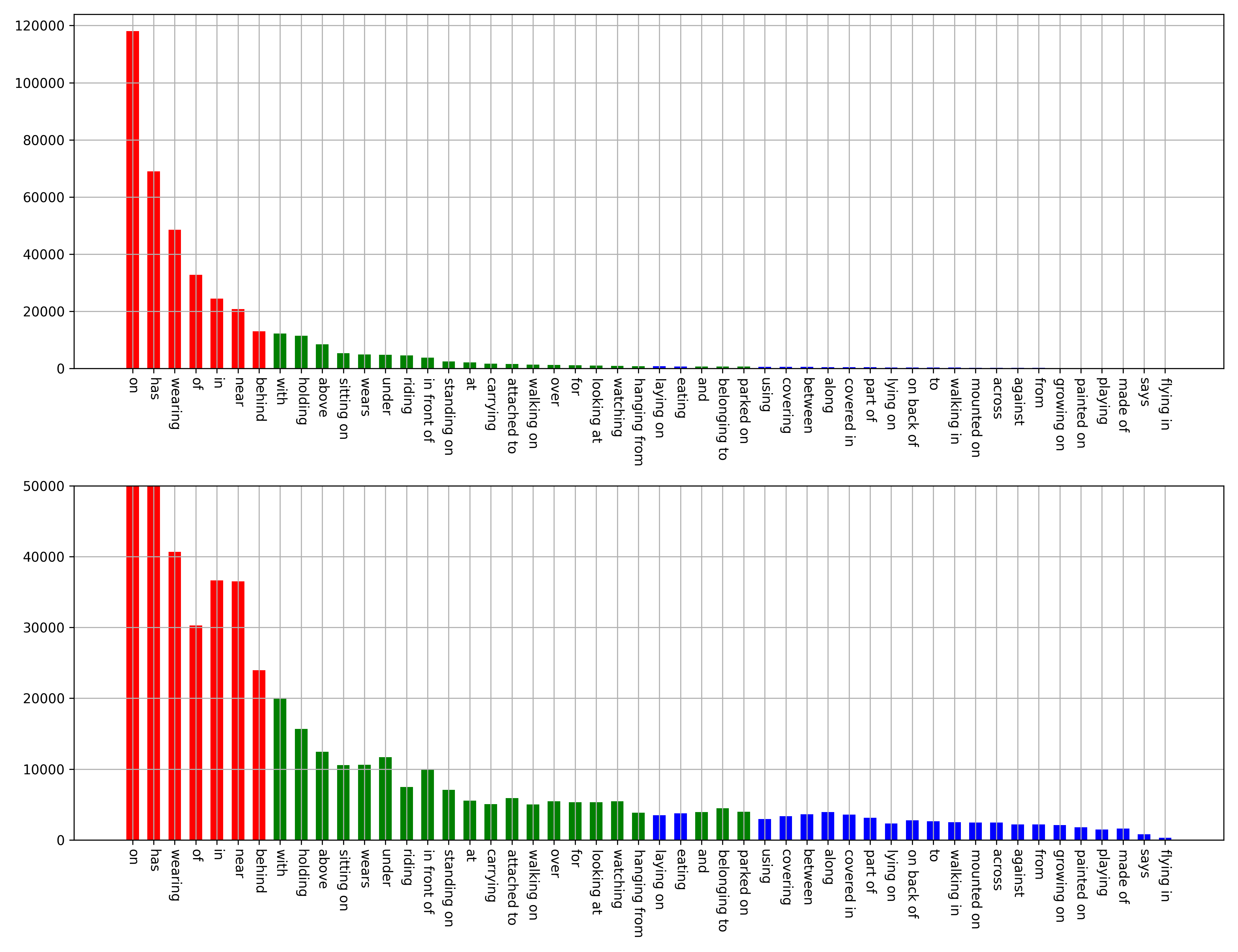}
\caption{ Three disjoint category groups in Visual Genome training split (follows a long-tail data distribution as demonstrated above):  $head$ (red bars), $body$ (green bars) and $tail$ (blue bars).  $y$ axis represents the number of samples.}
\label{fig_4}
\end{figure} 

\noindent \textbf{Evaluation Metrics:} Due to the reporting bias caused by the data imbalance \cite{tang2020unbiased}, in this paper, we choose mean Recall $mR@K$ as the evaluation metric instead of the traditional Recall $R@K$. Unlike $R@K$ which only concentrates on common predicate categories and underestimates the informative predicate categories, $mR@K$ averages the recalls across the predicate categories. Following the previous methods, we test the proposed NBP method on three tasks: predicate classification (PredCls), scene graph classification (SGCls) and scene graph detection (SGDet). Specifically, PredCls task aims to predict the predicate labels given the input image, the ground-truth bounding boxes and object labels; SGCls task tries to predict the object and predicate labels given the input image and the ground-truth bounding boxes; SGDet task generates the scene graph from the input image. 

\noindent \textbf{Implementation Details:} Following \cite{tang2020unbiased}, \cite{li2021bipartite}, in this experiment, ResNeXt-101-FPN \cite{he2016deep} (backbone) and Faster-RCNN \cite{ren2015faster} (object detector) are applied to construct the visual perception module. Like the previous methods, we choose a step training strategy, in which the parameters of the visual perception module are kept frozen during the training period and we only train the visual context reasoning module. The batch size $bs$ is set to 12. A bi-level data  resampling strategy \cite{li2021bipartite} is adopted in this experiment, in which we set the repeat factor $t=0.07$, instance drop rate $\gamma_{d}=0.7$ and weight of fusion the entities features $\rho=-5$. In this paper, we choose different numbers of NBP layers for the above three tasks. Specifically, we employ two NBP layers in the PredCls task and only one NBP layer is applied in the SGCls task. For the SGDet task, we use three NBP layers. An SGD optimizer with the learning rate of $0.008\times bs$ is applied to train the above three tasks.

\subsubsection{Comparison with State-of-the-art Methods}

For a fair comparison, in this experiment, we compare the proposed NBP method with several state-of-the-art baseline models. Some of them were reproduced using the author's latest codes, while others utilize the original codes but with a specific re-sampling strategy \cite{gupta2019lvis}. As demonstrated in Table 1, the proposed NBP method achieves the state-of-the-art performance in the SGCls and SGDet tasks, and comparable performance in PredCls task. Specifically, for the most representative SGDet task, compared with the previous best BGNN model, the proposed NBP method improves the $mR@50$ and $mR@100$ performance by $16.7\%$ and $20.6\%$.

Moreover, to investigate the biased relationship prediction problem caused by the long-tail data distribution,  we compare the $R@100$ performance on the long-tail category groups in the SGDet task in Table 1. For the informative $tail$ and $body$ predicate categories, the proposed NBP method  achieves the state-of-the-art performance. In particular, it outperforms the previous methods by a large margin for the most informative $tail$ predicate categories. This implies the proposed NBP method is capable of detecting the informative predicate categories, which are hindered by the problem of having a fewer samples for training. Compared with the previous SGG models, which predominantly detect the common predicate categories, it can achieve relatively unbiased training. To mitigate the biased relationship prediction problem caused by the long-tail data distribution, such unbiased training is much needed for SGG models.

Following \cite{guo2021general}, to achieve an even more unbiased training, we adopt a generic balance adjustment strategy in the proposed NBP method, aiming to correct two aspects of imbalance: the semantic space imbalance and the training sample imbalance. For the semantic space level imbalance, a semantic adjustment process is applied to induce the predictions by the NBP method to be more informative by constructing an appropriate transition matrix. For the training sample imbalance, a balanced predicate learning procedure is employed to extend the sampling space for informative predicates. Here, the term informative predicate is used in reference to  the Shannon information theory, in which the predicates occurring less frequently are deemed to contain more information. With such simple yet effective information measurement scheme, one can easily generate balanced training samples for the proposed NBP method.
\begin{table}[!t]
   \resizebox{\columnwidth}{!}{
   \begin{threeparttable}
	\renewcommand{\arraystretch}{1.5}
	\caption{Performance comparison on the Visual Genome dataset with a balance adjustment strategy.}
	\centering
    \begin{tabular}{@{\extracolsep{4pt}}*7c@{}}
	\toprule
	{} & \multicolumn{2}{c}{PredCls} & \multicolumn{2}{c}{SGCls} & \multicolumn{2}{c}{SGDet}\\ \cmidrule{2-3} \cmidrule{4-5} \cmidrule{6-7}
	{Method} & {mR@50} & {mR@100} & {mR@50} & {mR@100} & {mR@50} & {mR@100}\\
	\midrule
	Motifs+BA\cite{guo2021general}  & $29.7$ & $31.7$ & $16.5$ & $17.5$ & $13.5$ & $15.6$\\
	VCTree+BA\cite{guo2021general}  & $30.6$ & $32.6$ & $20.1$ & $21.2$ & $13.5$ & $15.7$\\
    Transformer+BA\cite{guo2021general}   & $31.9$ & $34.2$ & $18.5$ & $19.4$ & $14.8$ & $17.1$\\
    \textbf{NBP+BA} & $\mathbf{35.8}$ & $\mathbf{37.9}$ & $\mathbf{20.5}$ & $\mathbf{21.9}$ & $\mathbf{14.8}$ & $\mathbf{17.3}$\\
	\bottomrule
    \end{tabular}
    \begin{tablenotes}
	\item [\textbullet] Note: All the above methods apply the same balance adjustment strategy as in \cite{guo2021general} .
      \end{tablenotes}
    \end{threeparttable}
    }
\end{table} 

As demonstrated in Fig.5, compared with the NBP method, the resulting NBP+BA algorithm has more balanced training samples, in which the informative $tail$ and $body$  predicate categories have comparable training samples as the common $head$ predicate categories. For the informative $tail$ and $body$ predicate categories, the resulting NBP+BA algorithm generally outperforms the NBP method. In Fig.5, the black dots denote the $mR@100$ performances. It can be seen that the black dots for NBP+BA algorithm are generally higher than the ones for NBP method.

For a fair comparison, in this experiment, the resulting NBP+BA method is compared with three baseline models as presented in \cite{guo2021general}. Specifically, based on the Shannon information theory, the balanced predicate learning procedure discards the redundant training samples of the common $head$ group, and keeps most of the training samples of the informative $body$ and $tail$ groups. As a result, the training samples of the NBP+BA method are more balanced and the resulting data distribution is no longer a long-tail distribution. Moreover, with the transition matrix introduced in the semantic adjustment process, the predictions from the NBP method are further mapped into more informative ones. As demonstrated in Table 2, the resulting NBP+BA method outperforms the previous state-of-the-art methods by a large margin in Visual Genome benchmark, especially for the PredCls task.
\begin{figure}[!t]
\centering
\includegraphics[width=\linewidth]{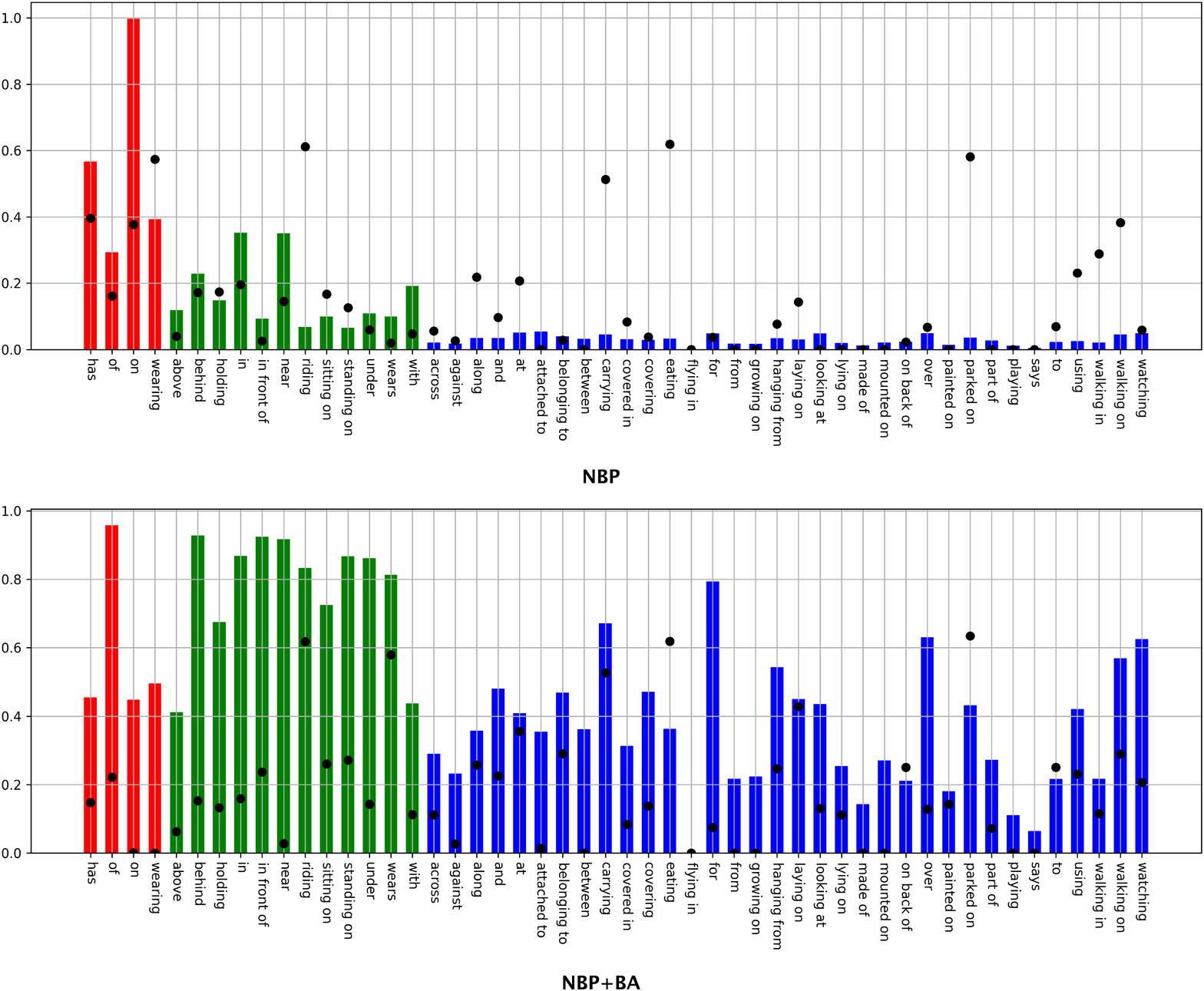}
\caption{Comparison of the $mR@100$ performance (represented as black dots) for each predicate category with the proposed NBP method and the resulting NBP+BA algorithm. Here, $y$ axis denotes the min-max normalized frequency. Compared with the NBP method, the resulting NBP+BA algorithm has more balanced training samples, in which the informative $tail$ (blue bars) and $body$ (green bars) predicate categories have comparable training samples as the common $head$ (red bars) predicate categories.}
\label{fig_5}
\end{figure} 

\subsection{Open Images V6}
\subsubsection{Experimental Configurations}
\textbf{Benchmark:} Open Images V6 \cite{alina2020open} is another popular SGG benchmark, which consists of 301 object categories and 31 predicate categories. Compared with Visual Genome, it provides a superior annotation quality. In Open Images V6, there are 126,368 training images, 5322 test images and 1813 validation images. In this experiment, we adopt the same data processing protocols as in \cite{lin2020gps}, \cite{zhang2019vrd}, \cite{alina2020open}.
\begin{table}[!t]
   \resizebox{\columnwidth}{!}{
   \begin{threeparttable}
	\renewcommand{\arraystretch}{1.5}
	\caption{A performance comparison on the Open Images V6 dataset.}
	\centering
    \begin{tabular}{@{\extracolsep{4pt}}*6c@{}}
	\toprule
	{Method} & {mR@50} & {R@50} & {wmAP\_rel} & {wmAP\_phr} & {score\_wtd} \\
	\midrule
	RelDN$^{ \dagger}$\cite{zhang2019vrd}  & $33.98$ & $73.08$ & $32.16$ & $33.39$ & $40.84$\\
	RelDN$^{\dagger*}$\cite{zhang2019vrd}  & $37.20$ & $75.34$ & $33.21$ & $34.31$ & $41.97$ \\
    VCTree$^{\dagger}$\cite{tang2019learning} & $33.91$ & $74.08$ & $34.16$ & $33.11$ & $40.21$ \\
    G-RCNN$^{\dagger}$\cite{yang2018graph}   & $34.04$ & $74.51$ & $33.15$ & $34.21$ & $41.84$ \\
	Motifs$^{\dagger}$\cite{zellers2018neural}  & $32.68$ & $71.63$ & $29.91$ & $31.59$ & $38.93$ \\
    VCTree-TDE$^{\dagger}$\cite{tang2020unbiased}  & $35.47$ & $69.30$ & $30.74$ & $32.80$ & $39.27$ \\
    GPS-Net$^{\dagger}$\cite{lin2020gps}   & $35.26$ & $74.81$ & $32.85$ & $33.98$ & $41.69$ \\
    GPS-Net$^{\dagger *}$\cite{lin2020gps}   & $38.93$ & $74.74$ & $32.77$ & $33.87$ & $41.60$ \\
    BGNN\cite{li2021bipartite}  & $40.45$ & $74.98$ & $33.51$ & $34.15$ & $42.06$ \\
    \textbf{NBP} & $\mathbf{41.97}$ & $\mathbf{75.54}$ & $\mathbf{34.44}$ & $\mathbf{35.66}$ & $\mathbf{43.08}$ \\
	\bottomrule
    \end{tabular}
    \begin{tablenotes}
	\item [\textbullet] Note: All the above methods apply ResNeXt-101-FPN as the backbone. $*$ means the re-sampling strategy \cite{gupta2019lvis} is applied in this method, and $\dagger$ depicts the reproduced results with the latest code from the authors. 
      \end{tablenotes}
    \end{threeparttable}
    }
\end{table} 

\noindent\textbf{Evaluation Metrics:} Following the evaluation protocols in \cite{lin2020gps}, \cite{zhang2019vrd}, \cite{alina2020open}, the following evaluation metrics are chosen in this experiment: the mean Recall$@50$ ($mR@50$), the regular Recall$@50$ ($R@50$), the weighted mean AP of relationships ($wmAP_{rel}$) and the weighted mean AP of phrase ($wmAP_{phr}$). Like \cite{lin2020gps}, \cite{alina2020open}, \cite{zhang2019vrd}, the weight metric score is defined as: $score_{wtd}=0.2\times R@50 + 0.4\times wmAP_{rel} + 0.4\times wmAP_{phr}$.
\begin{figure*}[!t]
\centering
\includegraphics[width=\linewidth]{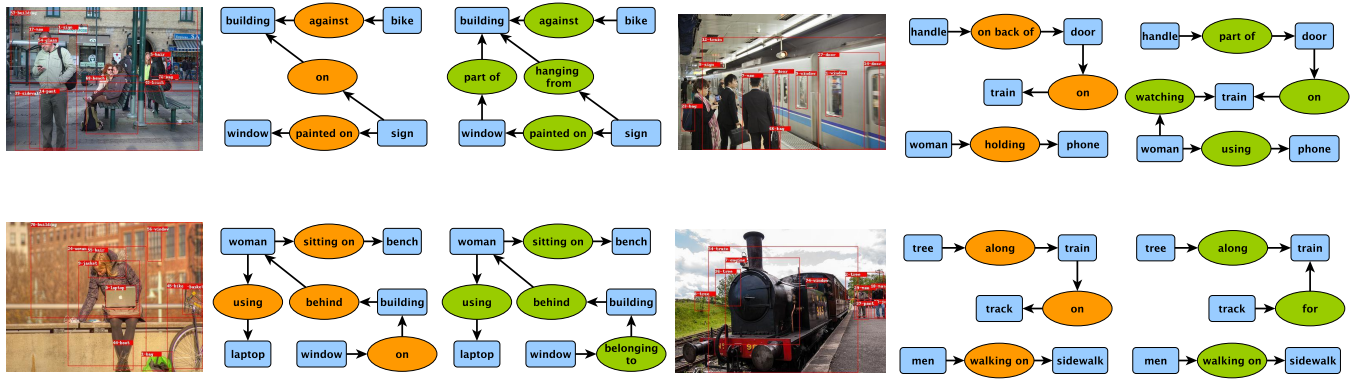}
\caption{ Visualization of the qualitative results of the proposed NBP (in orange) and NBP+BA (in green) methods. Due to the limited space, only the top predicates are shown in this image. The proposed NBP method is capable of detecting more informative $body$ and $tail$ predicates, i.e. $<painted \; on>$, $<sitting\; on>$, $<walking\; on>$ rather than the common $head$ predicate $on$. The resulting NBP+BA method further improves this capability, a demonstrated by detecting more informative predicates or even new ones.}
\label{fig_6}
\end{figure*} 

\noindent\textbf{Implementation Details:} Following the previous experiment, for the visual perception module, ResNeXt-101-FPN \cite{he2016deep} is employed as the backbone and Faster-RCNN \cite{ren2015faster} is applied as the object detector. As we adopt the step training strategy, freeze the model parameters of the above visual perception module and only train the visual context reasoning module. Moreover, the above bi-level data re-sampling strategy \cite{li2021bipartite} with the same settings is also utilized in this experiment. We set the batch size $bs$ to 12 and use two NBP layers in the visual context reasoning module. An Adam optimizer with learning rate of $0.0001$ is applied to train the proposed NBP method.

\subsubsection{Comparison with State-of-the-art Methods}

As demonstrated in Table 3, we compare the proposed NBP method with various state-of-the-art methods on the Open Images V6 benchmark. For a fair comparison, in this experiment, most of the baseline models are re-implemented with the author's latest codes. Some of them are used with an additional re-sampling strategy \cite{gupta2019lvis}. The proposed NBP method achieves the state-of-the-art performance on all evaluation metrics. Specifically, for the representative $mR@50$ metric, the proposed NBP method outperforms the previous methods by a large margin. Clearly, the above superior performance on complex Open Images V6 benchmark verifies the effectiveness of the proposed NBP method.

\subsection{Ablation Study}

In this section, for the proposed NBP method, we first investigate the impact of different types of aggregators on the final scene graph detection performance. Specifically, we compare the SGDet performances of two models: NBP method with the $max$ aggregator and NBP method with $mean$ aggregator, as shown in Table 4. We observe that the NBP method with the $mean$ aggregator constantly outperforms its counterpart NBP algorithm with the $max$ aggregator. This implies that, compared with the max-product method, the sum-product algorithm is more suitable for the scene graph generation task. As a result, in the proposed NBP architecture, we advocate the use of   the $mean$ aggregator, instead of the $max$ aggregator, as shown in Fig.3.
\begin{table}[!t]
   \resizebox{\columnwidth}{!}{
   \begin{threeparttable}
	\renewcommand{\arraystretch}{1.5}
	\caption{An ablation study of different types of aggregators.}
	\centering
    \begin{tabular}{@{\extracolsep{4pt}}*5c@{}}
	\toprule
	{Method} & {Aggregator Type} & {mR@20} & {mR@50} & {mR@100} \\
	\midrule
    NBP & $max$ & $9.2$ & $12.1$ & $14.2$  \\
    NBP & $\mathbf{mean}$ & $\mathbf{10.0}$ & $\mathbf{12.9}$ & $\mathbf{14.7}$  \\ 
	\bottomrule
    \end{tabular}
    \begin{tablenotes}
	\item [\textbullet] Note: Two types of aggregators - $max$ and $mean$ - are compared in this table.
      \end{tablenotes}
    \end{threeparttable}
    }
\end{table} 

Furthermore, we conduct another ablation study to investigate the impact of the proposed scoring function on the final scene graph generation performance. In this study, the baseline model is set to an NBP method with a scoring function containing only unary and pairwise interactions. As demonstrated in Table 5, the above baseline model is compared with the NBP method employing the proposed scoring function specified in Equation (12). It can be seen that the proposed scoring function with the higher order interactions generally produces better performance across all the evaluation metrics, which implies the global contextual information injected by the higher order interactions play an important role in generating a more consistent interpretation for an input image. This is mainly because the proposed scoring function reduces the model bias by achieving  a better bias-variance trade-off.
\begin{table}[!t]
   \resizebox{\columnwidth}{!}{
   \begin{threeparttable}
	\renewcommand{\arraystretch}{1.5}
	\caption{An ablation study of different types of scoring functions.}
	\centering
    \begin{tabular}{@{\extracolsep{4pt}}*5c@{}}
	\toprule
	{Method} & {Scoring Function} & {mR@20} & {mR@50} & {mR@100} \\
	\midrule
    NBP & without HO & $7.5$ & $10.5$ & $12.4$  \\
    NBP & \textbf{with HO} & $\mathbf{10.0}$ & $\mathbf{12.9}$ & $\mathbf{14.7}$  \\ 
	\bottomrule
    \end{tabular}
    \begin{tablenotes}
	\item [\textbullet] Note: HO stands for higher order.
      \end{tablenotes}
    \end{threeparttable}
    }
\end{table} 

\subsection{Visualization Results}

To intuitively demonstrate the superiority of the proposed method,  visualization of the qualitative results of the NBP and NBP+BA methods is presented in this section. As demonstrated in Fig.6, unlike the previous SGG models which mainly detect common $head$ predicates, the proposed NBP method is capable of detecting more informative $body$ and $tail$ predicates. For instance, rather than detecting the common predicate $on$, the NBP method generates more informative predicates like $<painted\; on>, $ $<sitting\; on>$, $<walking\; on>$. Moreover, the spatial informative predicates such as $<against>$ or $<on\; back\; of>$ can also be detected. Finally, the resulting NBP+BA method further improves the above capability. For example, for the top left image, the resulting NBP+BA method was able to detect more informative triplet $<sign\; hanging\; from\; building>$ rather than the common $<sign\; on\; building>$ triplet. It also detected a new triplet $<window\; part\; of\; building>$, which further enriches the resulting scene graph so that more structural information are available for the downstream computer vision tasks.
 
\section{Conclusion}

In this paper, we proposed a novel neural belief propagation method, which aims to solve two main drawbacks of the previous mean field message passing neural network models, namely that: 1) the output variables are considered to be fully independent within the approximation; 2) only pairwise interactions are incorporated into the associated scoring function. To find a better bias-variance trade-off, a novel scoring function incorporating higher order interactions is proposed. The proposed NBP method aims to simulate a classical sum-product algorithm to infer the optimum interpretations for an input image. We validated the proposed generic method on two popular scene graph generation benchmarks: Visual Genome and Open Images V6.  The extensive experimental results clearly demonstrate its superiority.

\ifCLASSOPTIONcompsoc
  \section*{Acknowledgments}
\else
  \section*{Acknowledgment}
\fi

This work was supported in part by the U.K. Defence Science and Technology Laboratory, and in part by the Engineering and Physical Research Council (collaboration between U.S. DOD, U.K. MOD, and U.K. EPSRC through the Multidisciplinary University Research Initiative) under Grant EP/R018456/1.

\ifCLASSOPTIONcaptionsoff
  \newpage
\fi

\bibliographystyle{IEEEtran}
\bibliography{IEEEabrv,Semantic}

\begin{thebibliography}{10}
\providecommand{\url}[1]{#1}
\csname url@samestyle\endcsname
\providecommand{\newblock}{\relax}
\providecommand{\bibinfo}[2]{#2}
\providecommand{\BIBentrySTDinterwordspacing}{\spaceskip=0pt\relax}
\providecommand{\BIBentryALTinterwordstretchfactor}{4}
\providecommand{\BIBentryALTinterwordspacing}{\spaceskip=\fontdimen2\font plus
\BIBentryALTinterwordstretchfactor\fontdimen3\font minus
  \fontdimen4\font\relax}
\providecommand{\BIBforeignlanguage}[2]{{%
\expandafter\ifx\csname l@#1\endcsname\relax
\typeout{** WARNING: IEEEtran.bst: No hyphenation pattern has been}%
\typeout{** loaded for the language `#1'. Using the pattern for}%
\typeout{** the default language instead.}%
\else
\language=\csname l@#1\endcsname
\fi
#2}}
\providecommand{\BIBdecl}{\relax}
\BIBdecl

\bibitem{sutton2006introduction}
C.~Sutton and A.~McCallum, ``An introduction to conditional random fields for
  relational learning,'' \emph{Introduction to statistical relational
  learning}, vol.~2, pp. 93--128, 2006.

\bibitem{zheng2015conditional}
S.~Zheng, S.~Jayasumana, B.~Romera-Paredes, V.~Vineet, Z.~Su, D.~Du, C.~Huang,
  and P.~H. Torr, ``Conditional random fields as recurrent neural networks,''
  in \emph{Proceedings of the IEEE international conference on computer
  vision}, 2015, pp. 1529--1537.

\bibitem{wainwright2008graphical}
M.~J. Wainwright, M.~I. Jordan \emph{et~al.}, ``Graphical models, exponential
  families, and variational inference,'' \emph{Foundations and Trends in
  Machine Learning}, vol.~1, no. 1--2, pp. 1--305, 2008.

\bibitem{fox2012tutorial}
C.~W. Fox and S.~J. Roberts, ``A tutorial on variational bayesian inference,''
  \emph{Artificial intelligence review}, vol.~38, no.~2, pp. 85--95, 2012.

\bibitem{murphy1999loopy}
K.~P. Murphy, Y.~Weiss, and M.~I. Jordan, ``Loopy belief propagation for
  approximate inference: an empirical study,'' in \emph{Proceedings of the
  Fifteenth conference on Uncertainty in artificial intelligence}, 1999, pp.
  467--475.

\bibitem{ihler2005loopy}
A.~T. Ihler, J.~W. Fisher~III, A.~S. Willsky, and D.~M. Chickering, ``Loopy
  belief propagation: convergence and effects of message errors.''
  \emph{Journal of Machine Learning Research}, vol.~6, no.~5, 2005.

\bibitem{liu2019visual}
D.~Liu, M.~Bober, and J.~Kittler, ``Visual semantic information pursuit: A
  survey,'' \emph{IEEE transactions on pattern analysis and machine
  intelligence}, vol.~43, no.~4, pp. 1404--1422, 2019.

\bibitem{zhang2020factor}
Z.~Zhang, F.~Wu, and W.~S. Lee, ``Factor graph neural networks,'' in
  \emph{Advances in Neural Information Processing Systems}, vol.~33, 2020, pp.
  8577--8587.

\bibitem{yedidia2003understanding}
J.~S. Yedidia, W.~T. Freeman, Y.~Weiss \emph{et~al.}, ``Understanding belief
  propagation and its generalizations,'' \emph{Exploring artificial
  intelligence in the new millennium}, vol.~8, pp. 236--239, 2003.

\bibitem{yedidia2005constructing}
J.~S. Yedidia, W.~T. Freeman, and Y.~Weiss, ``Constructing free-energy
  approximations and generalized belief propagation algorithms,'' \emph{IEEE
  Transactions on information theory}, vol.~51, no.~7, pp. 2282--2312, 2005.

\bibitem{kschischang2001factor}
F.~R. Kschischang, B.~J. Frey, and H.-A. Loeliger, ``Factor graphs and the
  sum-product algorithm,'' \emph{IEEE Transactions on information theory},
  vol.~47, no.~2, pp. 498--519, 2001.

\bibitem{loeliger2007factor}
H.-A. Loeliger, J.~Dauwels, J.~Hu, S.~Korl, L.~Ping, and F.~R. Kschischang,
  ``The factor graph approach to model-based signal processing,''
  \emph{Proceedings of the IEEE}, vol.~95, no.~6, pp. 1295--1322, 2007.

\bibitem{you2016image}
Q.~You, H.~Jin, Z.~Wang, C.~Fang, and J.~Luo, ``Image captioning with semantic
  attention,'' in \emph{Proceedings of the IEEE conference on computer vision
  and pattern recognition}, 2016, pp. 4651--4659.

\bibitem{rennie2017self}
S.~J. Rennie, E.~Marcheret, Y.~Mroueh, J.~Ross, and V.~Goel, ``Self-critical
  sequence training for image captioning,'' in \emph{Proceedings of the IEEE
  conference on computer vision and pattern recognition}, 2017, pp. 7008--7024.

\bibitem{yang2019auto}
X.~Yang, K.~Tang, H.~Zhang, and J.~Cai, ``Auto-encoding scene graphs for image
  captioning,'' in \emph{Proceedings of the IEEE/CVF Conference on Computer
  Vision and Pattern Recognition}, 2019, pp. 10\,685--10\,694.

\bibitem{teney2017graph}
D.~Teney, L.~Liu, and A.~van Den~Hengel, ``Graph-structured representations for
  visual question answering,'' in \emph{Proceedings of the IEEE conference on
  computer vision and pattern recognition}, 2017, pp. 1--9.

\bibitem{anderson2018bottom}
P.~Anderson, X.~He, C.~Buehler, D.~Teney, M.~Johnson, S.~Gould, and L.~Zhang,
  ``Bottom-up and top-down attention for image captioning and visual question
  answering,'' in \emph{Proceedings of the IEEE conference on computer vision
  and pattern recognition}, 2018, pp. 6077--6086.

\bibitem{shi2019explainable}
J.~Shi, H.~Zhang, and J.~Li, ``Explainable and explicit visual reasoning over
  scene graphs,'' in \emph{Proceedings of the IEEE/CVF Conference on Computer
  Vision and Pattern Recognition}, 2019, pp. 8376--8384.

\bibitem{dai2017detecting}
B.~Dai, Y.~Zhang, and D.~Lin, ``Detecting visual relationships with deep
  relational networks,'' in \emph{Proceedings of the IEEE conference on
  computer vision and Pattern recognition}, 2017, pp. 3076--3086.

\bibitem{li2018factorizable}
Y.~Li, W.~Ouyang, B.~Zhou, J.~Shi, C.~Zhang, and X.~Wang, ``Factorizable net:
  an efficient subgraph-based framework for scene graph generation,'' in
  \emph{Proceedings of the European Conference on Computer Vision (ECCV)},
  2018, pp. 335--351.

\bibitem{qi2019attentive}
M.~Qi, W.~Li, Z.~Yang, Y.~Wang, and J.~Luo, ``Attentive relational networks for
  mapping images to scene graphs,'' in \emph{Proceedings of the IEEE/CVF
  Conference on Computer Vision and Pattern Recognition}, 2019, pp. 3957--3966.

\bibitem{zellers2018neural}
R.~Zellers, M.~Yatskar, S.~Thomson, and Y.~Choi, ``Neural motifs: Scene graph
  parsing with global context,'' in \emph{Proceedings of the IEEE Conference on
  Computer Vision and Pattern Recognition}, 2018, pp. 5831--5840.

\bibitem{yang2018graph}
J.~Yang, J.~Lu, S.~Lee, D.~Batra, and D.~Parikh, ``Graph r-cnn for scene graph
  generation,'' in \emph{Proceedings of the European conference on computer
  vision (ECCV)}, 2018, pp. 670--685.

\bibitem{tang2019learning}
K.~Tang, H.~Zhang, B.~Wu, W.~Luo, and W.~Liu, ``Learning to compose dynamic
  tree structures for visual contexts,'' in \emph{Proceedings of the IEEE
  Conference on Computer Vision and Pattern Recognition}, 2019, pp. 6619--6628.

\bibitem{woo2018linknet}
S.~Woo, D.~Kim, D.~Cho, and I.~S. Kweon, ``Linknet: Relational embedding for
  scene graph,'' \emph{Advances in Neural Information Processing Systems},
  vol.~31, pp. 560--570, 2018.

\bibitem{lin2020gps}
X.~Lin, C.~Ding, J.~Zeng, and D.~Tao, ``Gps-net: Graph property sensing network
  for scene graph generation,'' in \emph{Proceedings of the IEEE/CVF Conference
  on Computer Vision and Pattern Recognition}, 2020, pp. 3746--3753.

\bibitem{chawla2002smote}
N.~V. Chawla, K.~W. Bowyer, L.~O. Hall, and W.~P. Kegelmeyer, ``Smote:
  synthetic minority over-sampling technique,'' \emph{Journal of artificial
  intelligence research}, vol.~16, pp. 321--357, 2002.

\bibitem{shen2016relay}
L.~Shen, Z.~Lin, and Q.~Huang, ``Relay backpropagation for effective learning
  of deep convolutional neural networks,'' in \emph{European conference on
  computer vision}.\hskip 1em plus 0.5em minus 0.4em\relax Springer, 2016, pp.
  467--482.

\bibitem{mahajan2018exploring}
D.~Mahajan, R.~Girshick, V.~Ramanathan, K.~He, M.~Paluri, Y.~Li, A.~Bharambe,
  and L.~Van Der~Maaten, ``Exploring the limits of weakly supervised
  pretraining,'' in \emph{Proceedings of the European conference on computer
  vision (ECCV)}, 2018, pp. 181--196.

\bibitem{gupta2019lvis}
A.~Gupta, P.~Dollar, and R.~Girshick, ``Lvis: A dataset for large vocabulary
  instance segmentation,'' in \emph{Proceedings of the IEEE/CVF Conference on
  Computer Vision and Pattern Recognition}, 2019, pp. 5356--5364.

\bibitem{hu2020learning}
X.~Hu, Y.~Jiang, K.~Tang, J.~Chen, C.~Miao, and H.~Zhang, ``Learning to segment
  the tail,'' in \emph{Proceedings of the IEEE/CVF Conference on Computer
  Vision and Pattern Recognition}, 2020, pp. 14\,045--14\,054.

\bibitem{li2021bipartite}
R.~Li, S.~Zhang, B.~Wan, and X.~He, ``Bipartite graph network with adaptive
  message passing for unbiased scene graph generation,'' in \emph{Proceedings
  of the IEEE/CVF Conference on Computer Vision and Pattern Recognition}, 2021,
  pp. 11\,109--11\,119.

\bibitem{cao2019learning}
K.~Cao, C.~Wei, A.~Gaidon, N.~Arechiga, and T.~Ma, ``Learning imbalanced
  datasets with label-distribution-aware margin loss,'' in \emph{Proceedings of
  the 33rd International Conference on Neural Information Processing Systems},
  2019, pp. 1567--1578.

\bibitem{cui2019class}
Y.~Cui, M.~Jia, T.-Y. Lin, Y.~Song, and S.~Belongie, ``Class-balanced loss
  based on effective number of samples,'' in \emph{Proceedings of the IEEE/CVF
  conference on computer vision and pattern recognition}, 2019, pp. 9268--9277.

\bibitem{gidaris2018dynamic}
S.~Gidaris and N.~Komodakis, ``Dynamic few-shot visual learning without
  forgetting,'' in \emph{Proceedings of the IEEE Conference on Computer Vision
  and Pattern Recognition}, 2018, pp. 4367--4375.

\bibitem{zhou2020bbn}
B.~Zhou, Q.~Cui, X.-S. Wei, and Z.-M. Chen, ``Bbn: Bilateral-branch network
  with cumulative learning for long-tailed visual recognition,'' in
  \emph{Proceedings of the IEEE/CVF Conference on Computer Vision and Pattern
  Recognition}, 2020, pp. 9719--9728.

\bibitem{guo2021general}
Y.~Guo, L.~Gao, X.~Wang, Y.~Hu, X.~Xu, X.~Lu, H.~T. Shen, and J.~Song, ``From
  general to specific: Informative scene graph generation via balance
  adjustment,'' in \emph{Proceedings of the IEEE/CVF International Conference
  on Computer Vision}, 2021, pp. 16\,383--16\,392.

\bibitem{tang2020unbiased}
K.~Tang, Y.~Niu, J.~Huang, J.~Shi, and H.~Zhang, ``Unbiased scene graph
  generation from biased training,'' in \emph{Proceedings of the IEEE/CVF
  Conference on Computer Vision and Pattern Recognition}, 2020, pp. 3716--3725.

\bibitem{kuck2020belief}
J.~Kuck, S.~Chakraborty, H.~Tang, R.~Luo, J.~Song, A.~Sabharwal, and S.~Ermon,
  ``Belief propagation neural networks,'' \emph{arXiv preprint
  arXiv:2007.00295}, 2020.

\bibitem{satorras2021neural}
V.~G. Satorras and M.~Welling, ``Neural enhanced belief propagation on factor
  graphs,'' in \emph{International Conference on Artificial Intelligence and
  Statistics}.\hskip 1em plus 0.5em minus 0.4em\relax PMLR, 2021, pp. 685--693.

\bibitem{meltzer2009convergent}
T.~Meltzer, A.~Globerson, and Y.~Weiss, ``Convergent message passing
  algorithms: a unifying view,'' in \emph{Proceedings of the Twenty-Fifth
  Conference on Uncertainty in Artificial Intelligence}, 2009, pp. 393--401.

\bibitem{gilmer2017neural}
J.~Gilmer, S.~S. Schoenholz, P.~F. Riley, O.~Vinyals, and G.~E. Dahl, ``Neural
  message passing for quantum chemistry,'' in \emph{International conference on
  machine learning}.\hskip 1em plus 0.5em minus 0.4em\relax PMLR, 2017, pp.
  1263--1272.

\bibitem{krishna2017visual}
R.~Krishna, Y.~Zhu, O.~Groth, J.~Johnson, K.~Hata, J.~Kravitz, S.~Chen,
  Y.~Kalantidis, L.-J. Li, D.~A. Shamma \emph{et~al.}, ``Visual genome:
  Connecting language and vision using crowdsourced dense image annotations,''
  \emph{International journal of computer vision}, vol. 123, no.~1, pp. 32--73,
  2017.

\bibitem{alina2020open}
A.~Kuznetsova, H.~Rom, N.~Alldrin, J.~Uijlings, I.~Krasin, J.~Pont-Tuset,
  S.~Kamali, S.~Popov, M.~Malloci, T.~Duerig, and V.~Ferrari, ``The open images
  dataset v4: Unified image classification, object detection, and visual
  relationship detection at scale,'' \emph{International journal of computer
  vision}, 2020.

\bibitem{zhang2019vrd}
J.~Zhang, K.~J. Shih, A.~Elgammal, A.~Tao, and B.~Catanzaro, ``Graphical
  contrastive losses for scene graph parsing,'' in \emph{Proceedings of the
  IEEE Conference on Computer Vision and Pattern Recognition}, 2019.

\bibitem{li2017scene}
Y.~Li, W.~Ouyang, B.~Zhou, K.~Wang, and X.~Wang, ``Scene graph generation from
  objects, phrases and region captions,'' in \emph{Proceedings of the IEEE
  international conference on computer vision}, 2017, pp. 1261--1270.

\bibitem{xu2017scene}
D.~Xu, Y.~Zhu, C.~B. Choy, and L.~Fei-Fei, ``Scene graph generation by
  iterative message passing,'' in \emph{Proceedings of the IEEE conference on
  computer vision and pattern recognition}, 2017, pp. 5410--5419.

\bibitem{liu2019large}
Z.~Liu, Z.~Miao, X.~Zhan, J.~Wang, B.~Gong, and S.~X. Yu, ``Large-scale
  long-tailed recognition in an open world,'' in \emph{Proceedings of the
  IEEE/CVF Conference on Computer Vision and Pattern Recognition}, 2019, pp.
  2537--2546.

\bibitem{he2016deep}
K.~He, X.~Zhang, S.~Ren, and J.~Sun, ``Deep residual learning for image
  recognition,'' in \emph{Proceedings of the IEEE conference on computer vision
  and pattern recognition}, 2016, pp. 770--778.

\bibitem{ren2015faster}
S.~Ren, K.~He, R.~Girshick, and J.~Sun, ``Faster r-cnn: Towards real-time
  object detection with region proposal networks,'' \emph{Advances in neural
  information processing systems}, vol.~28, pp. 91--99, 2015.

\end{thebibliography}

\end{document}